\documentclass[11pt]{article}

\usepackage[margin=1in]{geometry}
\usepackage{amsmath, amssymb, amsfonts}
\usepackage{booktabs}
\usepackage{graphicx}
\usepackage{hyperref}
\usepackage{url}

\title{Skill-Constrained Model Predictive Control for Resilient Manufacturing Supply Chains}
\author{Carlos Eduardo Sanoja\\[2pt]
  Quanta Labs, LLC\\
  Professor, FCEA, Universidad Monte\'avila\\
  Edificio Lomas del Sol, Calle Humboldt, Lomas del Sol, Caracas, Venezuela\\
  \texttt{csanoja@somosquanta.com}\\
  ORCID: \href{https://orcid.org/0009-0000-0339-7072}{0009-0000-0339-7072}}
\date{\today}

\begin{document}
\maketitle

\begin{abstract}
In skill-constrained production-inventory systems, the qualified human
capacity available tomorrow depends on training decisions made today:
production requires certified workers, certifications decay unless
maintained, and training consumes the same scarce worker hours that
production needs now. We study a closed-loop skill-constrained model
predictive controller that, at every shift, solves a finite-horizon
mixed-integer program over production, inventory, backlog, and training,
with binary predicted certification, hard production eligibility, and an
interpretable terminal value that prices certified-capacity gaps at the
horizon boundary; only the first-period action is applied before
replanning. On synthetic, seed-controlled SkillChain-Gym scenarios ---
announced and surprise new-skill shocks, demand shocks, absenteeism,
forecast- and availability-quality modes, capacity-boundary and
training-rate sweeps, and negative controls --- we evaluate the controller
against production-only and maintenance-only ablations, static
cross-training insurance plans, and a strong reactive heuristic, under an
ex-ante locked configuration and paired statistics. The result is regime
dependence, not superiority: no policy class dominates. Predictive control
helps when skill or labor bottlenecks are forecastable early enough for
training to complete; lean static insurance remains hard to beat under
surprise shocks, near the demand--capacity boundary, and wherever pre-shock
slack makes insurance cheap. Attribution ablations separate certification
maintenance, re-acquisition of lapsed certifications, and greenfield skill
acquisition. Forecastability, not adaptivity per se, decides when
predictive control pays.
\end{abstract}

\section{Introduction}
\label{sec:p2intro}

Production and service systems are usually planned as if labor capacity were
exogenous: workers appear as a fixed resource, and the planning question is
how to allocate materials, machines, and inventory around them. In
skill-constrained operations this abstraction fails in a specific way ---
the qualified capacity available tomorrow depends on training and
certification decisions made today. Workforce planning research has long
recognized skills, cross-training, and learning as first-class modeling
objects \cite{deBruecker2015workforceSkills,vanDenBergh2013personnelScheduling},
and workforce reconfiguration is increasingly treated as a resilience lever
in manufacturing \cite{hashemiPetroodi2021workforceReconfiguration,
prunet2024humanAwareReview}. When production requires certified workers and
certifications are dynamic, qualified human capacity --- not machines or
materials --- can be the binding operational resource. Reskilling then
stops being a background human-resources decision and becomes a control
action.

What makes the resulting control problem hard is a tight intertemporal
coupling. Training consumes the same scarce worker hours that production
needs now, so building future capability always costs current output
\cite{heuser2022flexibleBudgetedTraining,kafiabad2020onJobTraining}.
Skills decay when unused, so certification is a maintained asset rather
than a one-off purchase
\cite{kher1999learningForgetting,biskup2008learningReview,nembhard2012parallelLearningForgetting}.
Disruptions interact with both mechanisms: demand spikes and absences stress
existing certified capacity, while new-product introductions can require a
skill that no worker currently holds. Whether such shocks are forecast,
announced late, or hidden until onset changes the problem fundamentally,
because training has a lag --- reacting after a shock can be structurally
too late when post-shock demand leaves no slack for diverting hours into
training.

Each ingredient of this problem is well studied in isolation. Model
predictive and receding-horizon control of inventories, production, and
supply chains is a mature field
\cite{braun2003mpcDemandNetworks,pereaLopez2003mpcSupplyChain,
doganis2008mpcForecasting,li2009robustSupplyChainMpc,schildbach2016scenarioMpc},
but it treats labor as exogenous or absent. Workforce planning with skills,
training, learning, and forgetting is likewise mature
\cite{deBruecker2015workforceSkills,ruf2022hierarchicalSkills,
valeva2017balancingFlexibilityInventory}, but these models are
predominantly open-loop planning formulations rather than closed-loop
controllers. The closest control-side work models human activity-time
uncertainty or machine-capability ``skills'' inside MPC
\cite{ruppert2020fuzzyOpenStationMpc,wenzelburger2021flexibleJobShopMpc}
without worker skill evolution or training actions. We therefore make no
novelty claim for supply-chain MPC or for workforce-training models; the
contribution is their closed-loop combination: a receding-horizon controller
in which worker skill levels are observed dynamic states and training is an
online, capacity-consuming control action coupled to inventory and backlog
dynamics.

Concretely, we formulate skill-constrained production-inventory control on
the SkillChain-Gym benchmark of our companion paper
(Section~\ref{sec:p2formulation}): continuous skill levels with hard
threshold certification, production eligibility requiring certification,
geometric forgetting, and a shared per-worker time budget for production and
training. The controller (Section~\ref{sec:method}) solves, at every shift,
a finite-horizon mixed-integer program with binary predicted certification
and an interpretable terminal skill-bottleneck value that prices certified
capacity gaps left open at the horizon boundary, applies only the
first-period action, and replans. Attribution ablations isolate what the
skill machinery is worth: a production-only controller, a maintenance-only
controller that can preserve but never acquire certifications, and the full
controller with and without the terminal value, evaluated against
deliberately favorable static cross-training plans and a strong reactive
heuristic under an ex-ante locked primary configuration and paired
statistics (Section~\ref{sec:p2setup}).

The empirical picture (Section~\ref{sec:p2results}) is a regime map, not a
ranking. No policy class dominates. Predictive control helps when skill or
labor bottlenecks are forecastable early enough for training to complete:
forecast-visible new-skill bottlenecks, announced demand shocks where
inventory anticipation combines with certification maintenance, announced
absence windows, and slow-training regimes where the terminal value prevents
short-horizon controllers from ignoring visible future gaps. Lean static
insurance remains hard to beat when shocks are hidden, when reaction
transients are structurally unrecoverable near the demand--capacity
boundary, and wherever pre-shock slack makes insurance cheap. Degrading
forecast quality degrades the controller in an interpretable order,
making explicit that its advantages are forecast leverage rather than
generic adaptivity.

Our contributions are:
\begin{enumerate}
  \item a closed-loop skill-constrained MPC formulation for single-site
    production-inventory systems with dynamic worker skill states, hard
    certification, forgetting, and capacity-consuming training;
  \item a mixed-integer implementation with an interpretable terminal
    skill-bottleneck value and a variant/ablation chain separating
    production-only control, certification maintenance, no-terminal MPC,
    and full skill-aware MPC;
  \item a reproducible, deterministic experiment suite on the SkillChain-Gym
    simulator covering announced and surprise new-skill shocks, demand
    shocks, absenteeism, forecast and availability-forecast quality,
    capacity-boundary sweeps, training-rate sensitivity, and negative
    controls; and
  \item a mechanism-level regime analysis that separates certification
    maintenance, lapsed-certification re-acquisition, and greenfield skill
    acquisition with within-episode certification-event counters, and shows
    that forecastability --- of demand, of new skill requirements, and of
    labor availability --- decides when predictive control outperforms
    static insurance.
\end{enumerate}

\section{Related Work}

\subsection{Model predictive control for production-inventory and supply-chain systems}

Model predictive control (MPC) has a long history in production-inventory and supply-chain planning. Early work showed how receding-horizon optimization can manage multi-product, multi-echelon demand networks and supply-chain profit objectives under capacity, storage, production, and shipment constraints \cite{braun2003mpcDemandNetworks,braun2003semiconductorDemandNetworks,pereaLopez2003mpcSupplyChain}. Reviews of control-theoretic production-inventory and supply-chain models further establish that inventory, backlog, demand amplification, ordering, and material-flow dynamics have been studied extensively through control methods \cite{ortega2004controlProductionInventory,sarimveis2008dynamicSupplyChains}. Subsequent work developed centralized, robust, scenario-based, and forecasting-enhanced predictive controllers for supply-chain and inventory systems \cite{li2009robustSupplyChainMpc,fu2014centralisedSupplyChainMpc,fu2015epsacInventory,schildbach2016scenarioMpc,doganis2008mpcForecasting,alvarezRodriguez2017biomassMpc,hipolito2020continuousFlowMpc}. Recent receding-horizon game formulations also model competitive supply chains under demand spikes and supply shocks \cite{hall2024recedingHorizonGamesSupplyChain}. These papers motivate our control formulation, but generally treat productive capacity as exogenous or machine/process constrained rather than as a dynamic function of workforce skill development.

MPC has also been applied directly to manufacturing scheduling and shop-floor control. Examples include multilayer MPC for semiconductor lines, fab-wide scheduling, multi-product job-shop control, and flexible job-shop scheduling \cite{vargasVillamil2000multilayerMpc,jang2013fabWideMpc,sprodowski2020jobShopMpc,wenzelburger2021flexibleJobShopMpc}. The flexible job-shop work of Wenzelburger and Allgower is especially close in terminology because it models task and manufacturing-unit ``skills'' within an MPC scheduling framework \cite{wenzelburger2021flexibleJobShopMpc}; however, those skills describe machine or manufacturing-unit capabilities, not worker competencies that evolve through training. Ruppert et al. incorporate uncertain operator activity times in an MPC controller for manual assembly lines \cite{ruppert2020fuzzyOpenStationMpc}, but do not model worker skill acquisition or training as a control action.

\subsection{Workforce planning with skills, learning, and training}

The operations-research literature on workforce planning already contains extensive models of skills, skill levels, multi-skilling, learning, forgetting, cross-training, and human factors. Broad reviews cover personnel scheduling, workforce planning with skills, workforce reconfiguration in manufacturing, multi-skilling in scheduling, and human-aware logistics/manufacturing optimization \cite{vanDenBergh2013personnelScheduling,deBruecker2015workforceSkills,hashemiPetroodi2021workforceReconfiguration,afsharNadjafi2021multiskillingReview,prunet2024humanAwareReview,prunet2024humanAwareModels,catalano2025mannedAssembly}. Learning and forgetting in worker assignment and scheduling have also been modeled for dual-resource systems, parallel systems, cellular manufacturing, job rotation, and workforce assignment \cite{kher1999learningForgetting,zamiska2007workerDeployment,biskup2008learningReview,nembhard2012parallelLearningForgetting,azizi2010jobRotation,chu2019workerAssignmentLearning}. Integer-programming reformulations make nonlinear worker learning curves tractable in assignment and planning models \cite{hewitt2015ipLearning,jin2016ipLearningMakespan}, and learning-curve selection has been studied empirically in production economics \cite{grosse2015learningCurve}.

Several papers are particularly important constraints on our novelty claim. Azizi and Liang jointly optimize worker assignment, flexibility acquisition, task rotation, and training schedules in manufacturing \cite{aziziLiang2013workerAssignment}. Valeva et al. study workforce planning with learning, stochastic demand, and inventory as a flexibility buffer \cite{valeva2017balancingFlexibilityInventory,valeva2017matheuristicLearning}. Cavagnini et al. extend this line by modeling uncertain learning and forgetting rates with assignment, cross-training, and practice decisions \cite{cavagnini2020uncertainLearningRates}. Heuser et al. explicitly study production workforce planning with flexible or budgeted training, volatile demand, learning-by-doing, and forgetting, where training consumes capacity that could otherwise be used for production \cite{heuser2022flexibleBudgetedTraining}. Ruf et al. formulate hierarchical skills, long-term training, and random resignations as a multistage workforce capacity planning problem using approximate dynamic programming \cite{ruf2022hierarchicalSkills}. Henao et al. model multiskilled personnel assignment with learning-forgetting dynamics and k-chaining policies \cite{henao2023kChaining}, with related benchmark data for uncertain multiskilled personnel demand \cite{henao2024benchmarkDataset}.

\subsection{Integrated operations and training}

Integrated planning of operations and training is not new. De Bruecker et al. optimize aircraft maintenance skill mix and training schedules using a three-stage mixed-integer approach in which training affects workforce availability \cite{deBruecker2018aircraftTraining}. Kafiabad et al. integrate procurement, production, inventory, and on-job training in maintenance logistics networks \cite{kafiabad2020onJobTraining}, and later study workforce training and operations planning for maintenance centers under demand uncertainty \cite{kafiabad2022demandUncertainty}. These are the closest prior works to the present topic because they already connect operations, inventory-related decisions, certified operators, and training. Their focus, however, is tactical maintenance planning through deterministic or stochastic mathematical programming, not closed-loop MPC over observed inventory, backlog, availability, and worker skill states.

\subsection{Positioning of this paper}

The contribution of this paper is therefore not the use of MPC in supply chains, nor the modeling of worker learning or training in isolation. Instead, we study a closed-loop predictive-control formulation in which worker skills are part of the system state and training/reskilling is an explicit capacity-consuming control action coupled to production and inventory/backlog dynamics. This distinguishes the proposed controller from supply-chain MPC models that omit workforce skill evolution, and from workforce-training models that optimize over a fixed planning horizon without repeated state observation, forecast updates, and receding-horizon response to disruptions.

\section{Problem Formulation}
\label{sec:p2formulation}

We formulate the control problem on the single-site production-inventory
system with stylized worker skill-state dynamics defined by the SkillChain-Gym
benchmark. The scope is deliberately narrow: one site, multiple products, a
small set of skills, one aggregate production-capacity pool, and
inventory/backlog dynamics. Procurement, supplier delays, multi-echelon
flows, job-shop routing, soft (graded) productivity, and learning-by-doing
are excluded by design, and all instances are synthetic and seed-controlled;
no real data are used. This section defines the plant, the admissible
actions, and the control objective; Section~\ref{sec:method} describes the
receding-horizon implementation.

\paragraph{Sets and parameters.}
Shifts $t \in \{0, \dots, T-1\}$ (one period is one work shift; $T = 60$),
products $p \in \mathcal{P}$, workers $w \in \mathcal{W}$, and skills
$k \in \mathcal{K}$, with one primary skill $k(p)$ per product. Static
parameters: productivity $v_p > 0$ (units per certified worker-hour),
nominal worker hours $H_w$ per shift, certification thresholds
$\theta_k \in [0,1]$, training gain $\alpha^{\mathrm{train}} \ge 0$ per
training hour, forgetting rates $\delta_k \ge 0$ per shift, training-seat
capacities $cap^{\mathrm{train}}_k$, aggregate production-capacity hours
$C_t$, and cost coefficients $c^B_p$ (backlog), $c^I_p$ (holding), $c^q_p$
(production), and $c^Y$ (training) per unit and shift.

\paragraph{State.}
At shift $t$ the state is
$\bigl(I_t, B_t, \hat{D}_t, A_t, C_t, S_t, Q_t\bigr)$:
inventory $I_{p,t} \ge 0$, backlog $B_{p,t} \ge 0$, a demand-forecast window
$\hat{D}_{p,t:t+F}$, worker availability $A_{w,t} \in [0, H_w]$, the capacity
pool, and continuous skill levels $S_{w,k,t} \in [0,1]$ from which
certification is derived by hard thresholding,
\begin{equation}
  Q_{w,k,t} \;=\; \mathbf{1}\!\left[S_{w,k,t} \ge \theta_k\right].
  \label{eq:pf-cert}
\end{equation}
There is no partial productivity below the threshold, no separate
training-progress state, and no skill growth from production work.

\paragraph{Actions and feasibility.}
The action $u_t = (a^{\mathrm{prod}}_t, a^{\mathrm{train}}_t)$ allocates
nonnegative worker-hours to production ($a^{\mathrm{prod}}_{w,p,t}$) and to
training ($a^{\mathrm{train}}_{w,k,t}$), subject to
\begin{align}
  \sum_p a^{\mathrm{prod}}_{w,p,t} + \sum_k a^{\mathrm{train}}_{w,k,t}
    &\;\le\; A_{w,t} \quad \forall w,
  \label{eq:pf-budget} \\
  a^{\mathrm{prod}}_{w,p,t} = 0 \;\text{ whenever }\; Q_{w,k(p),t} = 0,
  \qquad
  \sum_{w,p} a^{\mathrm{prod}}_{w,p,t} \;\le\; C_t,
  \qquad
  \sum_w a^{\mathrm{train}}_{w,k,t} &\;\le\; cap^{\mathrm{train}}_k .
  \label{eq:pf-elig}
\end{align}
Constraint~\eqref{eq:pf-budget} is the central mechanism: training consumes
the same scarce worker time as production, so reskilling is never free.
Production output is
$q_{p,t} = v_p \sum_w a^{\mathrm{prod}}_{w,p,t}\, Q_{w,k(p),t}$.

\paragraph{Dynamics.}
Demand $D_{p,t}$ is realized after the action as the forecast mean plus
seed-controlled noise. Shipments serve current demand plus backlog from
inventory and current production, and skills decay geometrically while
growing linearly in training time:
\begin{align}
  \mathrm{ship}_{p,t} &= \min\bigl(I_{p,t} + q_{p,t},\, D_{p,t} + B_{p,t}\bigr),
  \quad
  I_{p,t+1} = I_{p,t} + q_{p,t} - \mathrm{ship}_{p,t},
  \quad
  B_{p,t+1} = B_{p,t} + D_{p,t} - \mathrm{ship}_{p,t},
  \label{eq:pf-ship} \\
  S_{w,k,t+1} &= \Pi_{[0,1]}\!\Bigl((1-\delta_k)\, S_{w,k,t}
    + \alpha^{\mathrm{train}} a^{\mathrm{train}}_{w,k,t}\Bigr),
  \qquad
  Q_{w,k,t+1} = \mathbf{1}\!\left[S_{w,k,t+1} \ge \theta_k\right].
  \label{eq:pf-skill}
\end{align}
Forgetting makes certification a maintained asset rather than a one-off
purchase: an unmaintained skill eventually falls below threshold, and a
lapsed certification can only be recovered by further training.

\paragraph{Objective and information structure.}
The per-shift cost is
\begin{equation}
  c_t \;=\; \sum_p c^B_p B_{p,t+1} + \sum_p c^I_p I_{p,t+1}
    + \sum_p c^q_p q_{p,t} + c^Y \sum_{w,k} a^{\mathrm{train}}_{w,k,t},
  \label{eq:pf-cost}
\end{equation}
and the control problem is to choose a causal policy
$u_t = \pi(\cdot)$ minimizing the episode cost $\sum_{t=0}^{T-1} c_t$
under the scenario's disruption process (demand spikes, absenteeism windows,
and new-product activations that require an initially uncertified skill).
The policy observes only the environment's observation: in particular, the
demand-forecast window reflects each scenario's visibility semantics ---
announced events appear in the window before they occur, while surprise
activations are hidden until onset --- so anticipation is possible exactly
when, and only when, the scenario makes the future visible. The
intertemporal tension that makes this problem non-trivial is that serving
demand today and being \emph{able} to serve demand tomorrow draw on the same
worker hours through~\eqref{eq:pf-budget}, while
certification~\eqref{eq:pf-cert} makes future capacity a discrete,
training-lagged consequence of today's allocation. The receding-horizon
controller of Section~\ref{sec:method} approximates this problem with a
finite-horizon mixed-integer program whose prediction model convexifies the
shipment dynamics into a net inventory--backlog balance and represents
predicted certification with binary variables.

\section{Skill-Constrained Model Predictive Control}
\label{sec:method}

We study closed-loop receding-horizon control of a single-site
production-inventory system in which worker skills are dynamic states and
training is a control action that consumes the same worker hours as
production. The plant is the SkillChain-Gym environment of the companion
benchmark paper: continuous skill levels $S_{w,k,t} \in [0,1]$ per worker $w$
and skill $k$, hard threshold certification
$Q_{w,k,t} = \mathbf{1}[S_{w,k,t} \ge \theta_k]$, production eligibility
requiring certification, geometric forgetting at rate $\delta_k$, linear
training gain $\alpha^{\mathrm{train}}$ per hour, per-product primary skills
$k(p)$, productivity $v_p$, an aggregate production-capacity pool, and
inventory/backlog dynamics with stage costs on backlog ($c^B_p$), holding
($c^I_p$), production ($c^q_p$), and training ($c^Y$). No soft productivity
and no learning-by-doing are modeled.

\subsection{Closed-loop algorithm}
\label{sec:closedloop}

At every shift $t$ the controller (i)~observes the current state
$(I_t, B_t, A_t, C_t, S_t, Q_t)$ and the demand forecast window exposed by
the environment; (ii)~builds horizon forecasts $\hat{D}_{p,h|t}$,
$\hat{A}_{w,h|t}$, and $\hat{C}_{h|t}$ for $h = 0,\dots,H-1$;
(iii)~solves the finite-horizon mixed-integer program below;
(iv)~applies \emph{only} the first-period production and training hours
$(x_{w,p,0},\, y_{w,k,0})$; the environment then realizes demand and
disruptions and updates inventory, backlog, skills, and certifications; and
(v)~repeats from the updated state. Every controller in this paper replans at
every shift from the fresh observation; no plan is cached across steps.
Demand forecasts are read from the environment's observation window, which
reflects each scenario's visibility semantics (announced shocks appear in the
window ahead of time; surprise activations are hidden until onset), so the
controller never receives information the observation does not contain.
Availability and capacity forecasts default to persistence of the current
observation; alternative forecast-quality and availability-forecast modes are
controller-side transformations described in Section~\ref{sec:p2setup}.

\subsection{Finite-horizon mixed-integer program}
\label{sec:milp}

For horizon steps $h = 0,\dots,H-1$ the decision variables are production
hours $x_{w,p,h} \ge 0$, training hours $y_{w,k,h} \ge 0$, predicted
inventory $I_{p,h+1} \ge 0$ and backlog $B_{p,h+1} \ge 0$, predicted skills
$S_{w,k,h+1} \in [0,1]$, and binary certifications
$c_{w,k,h} \in \{0,1\}$ for $h \ge 1$, plus terminal variables introduced
below. The constraints are:
\begin{align}
  \sum_{p} x_{w,p,h} + \sum_{k} y_{w,k,h} &\;\le\; \hat{A}_{w,h|t}
  && \forall w, h
  \label{eq:p2budget} \\
  \sum_{w,p} x_{w,p,h} &\;\le\; \hat{C}_{h|t}
  && \forall h
  \label{eq:p2cap} \\
  \sum_{w} y_{w,k,h} &\;\le\; cap^{\mathrm{train}}_{k}
  && \forall k, h
  \label{eq:p2seats} \\
  x_{w,p,0} &\;\le\; \hat{A}_{w,0|t}\, Q_{w,k(p),t}
  && \forall w, p
  \label{eq:p2elig0} \\
  x_{w,p,h} &\;\le\; \hat{A}_{w,h|t}\, c_{w,k(p),h}
  && \forall w, p,\ h \ge 1
  \label{eq:p2elig} \\
  \theta_k\, c_{w,k,h} &\;\le\; S_{w,k,h}
  && \forall w, k,\ h \ge 1
  \label{eq:p2cert} \\
  I_{p,h+1} - B_{p,h+1} &\;=\; I_{p,h} - B_{p,h}
    + v_p \textstyle\sum_{w} x_{w,p,h} - \hat{D}_{p,h|t}
  && \forall p, h
  \label{eq:p2balance} \\
  S_{w,k,h+1} &\;\le\; (1-\delta_k)\, S_{w,k,h}
    + \alpha^{\mathrm{train}} y_{w,k,h}
  && \forall w, k, h
  \label{eq:p2skill}
\end{align}
with $S_{w,k,0}$, $I_{p,0}$, $B_{p,0}$ fixed to observed values.
Constraint~\eqref{eq:p2budget} is the capacity-consuming training mechanism:
production and training share each worker's time budget. First-period
eligibility~\eqref{eq:p2elig0} uses the \emph{observed} certification
$Q_{w,k(p),t}$, so the executed action is exactly feasible in the
environment; predicted eligibility~\eqref{eq:p2elig}--\eqref{eq:p2cert} uses
binary certification coupled to predicted skill, so the optimizer can plan to
train a worker and use the resulting capacity later in the horizon. The
skill dynamics enter as an inequality with the $[0,1]$ upper bound playing
the role of the saturation projection; because predicted skill only relaxes
constraints, the inequality binds wherever skill has value. The net
balance~\eqref{eq:p2balance} permits simultaneous predicted inventory and
backlog in principle, but with strictly positive holding costs the optimum
always drives $\min(I,B)$ to zero; the implementation asserts $c^I_p > 0$.

The stage objective sums backlog, holding, production, and training costs,
\begin{equation}
  J_t \;=\; \sum_{h=0}^{H-1} \Bigl[
    \sum_p c^B_p B_{p,h+1} + \sum_p c^I_p I_{p,h+1}
    + \sum_p c^q_p\, v_p \textstyle\sum_w x_{w,p,h}
    + c^Y (1 + \varepsilon h) \textstyle\sum_{w,k} y_{w,k,h}
  \Bigr] + V_f ,
  \label{eq:p2obj}
\end{equation}
where $\varepsilon = 10^{-3}$ is a tie-break that prefers earlier training
when timing is otherwise cost-degenerate; it is $0.1\%$ per step and does not
overcome the forgetting-driven preference for just-in-time training, nor does
it change any cost comparison reported here.

\subsection{Terminal skill-bottleneck value}
\label{sec:terminal}

The terminal value $V_f$ prices skill bottlenecks left open at the end of the
horizon. With binary terminal certifications $c^{T}_{w,k} \in \{0,1\}$
satisfying $\theta_k c^{T}_{w,k} \le S_{w,k,H}$, define the certified
capacity and the gap
\begin{equation}
  \mathrm{cap}_k \;=\; \sum_w \hat{A}_{w,H-1|t}\, c^{T}_{w,k},
  \qquad
  gap_k \;=\; \max\!\bigl(0,\; \widehat{sd}_k - \mathrm{cap}_k\bigr),
  \qquad
  V_f \;=\; \lambda_{\mathrm{gap}} \sum_k gap_k ,
  \label{eq:p2gap}
\end{equation}
where the forecasted skill demand $\widehat{sd}_k$ is the per-skill
\emph{maximum} of forecast demand hours from step $H$ to the end of the
visible forecast window. This choice makes the terminal value a cost-to-go
proxy: demand that is already visible beyond the horizon is exactly what the
truncated objective would otherwise ignore. Hidden (surprise) activations
contribute nothing to $\widehat{sd}_k$ until the environment reveals them, so
the terminal value cannot leak unannounced information. We evaluate
$\lambda_{\mathrm{gap}} \in \{0, 25, 100\}$; the penalty is interpretable (a
price per uncovered certified hour at the horizon boundary) and is ablated
rather than tuned --- the same three values are used in every scenario and
training-rate regime.

\subsection{Controller variants and baselines}
\label{sec:variants}

\begin{description}
  \item[ProductionOnlyMPC.] The receding-horizon production/inventory
    program without training variables or skill dynamics; eligibility over
    the whole horizon uses the currently observed certifications. It
    isolates the value of receding-horizon inventory planning alone.
  \item[MaintenanceMPC (attribution ablation).] The full program with
    training restricted to skills the worker currently holds: $y_{w,k,h}$ is
    fixed to zero whenever $Q_{w,k,t} = 0$. The controller can maintain
    currently held certifications against forgetting but can never acquire a
    new skill, and --- because the restriction re-reads the observed
    certification at every replan --- it can never recover a certification
    once it lapses. The gap between ProductionOnlyMPC and MaintenanceMPC
    therefore measures certification-maintenance value, and the gap between
    MaintenanceMPC and the full controller measures the value of acquiring
    or re-acquiring certifications.
  \item[SkillMPCNoTerminal.] The full program with
    $\lambda_{\mathrm{gap}} = 0$.
  \item[SkillMPCWithTerminal / Primary.] The full program with the terminal
    penalty. The \emph{primary configuration} is
    $\lambda_{\mathrm{gap}} = 25$, $H = 10$, fixed ex ante before any
    validation run; all horizon and $\lambda_{\mathrm{gap}}$ sweeps are
    sensitivity analyses, never best-of selection.
  \item[Static insurance baselines.] The open-loop cross-training plans of
    the benchmark paper: Static40 and Static60 at the default training rate,
    and rate-recalibrated variants (Static80 at the moderate rate;
    StaticSlow160, the same plan stretched to ten shifts, at the slow rate)
    so that every static plan certifies all its trained cells at the
    training rate it faces.
  \item[WaterFillingSkillGap.] The strongest reactive heuristic from the
    benchmark paper: skill-gap-driven training with proportional
    (water-filling) production allocation.
\end{description}

\subsection{Solver implementation and diagnostics}
\label{sec:solver}

Each replanning step solves the MILP with \texttt{scipy.optimize.milp}
(HiGHS branch and bound) with binary certification variables. We initially
evaluated a pure-LP relaxation with a continuous eligibility credit
$z \le S/\theta$ and rejected it: it grants uncertified workers large
phantom capacity in prediction, which suppresses training incentives and
makes the terminal gap vacuous, while a tighter ramp credit is non-convex.
The MILP is the honest model and remains fast at benchmark scale (typically
$3$--$120$\,ms per solve; the largest instances, $H{=}15$ or near-infeasible
demand, stay below $0.6$\,s). Every solve records its status and wall-clock
time; if a solve ever failed, the step would fall back to the water-filling
heuristic and be counted --- the fallback is a safety net, not part of the
method, and it never triggered in any accepted run. The executed
first-period action is exactly feasible by construction
(Eq.~\ref{eq:p2elig0}), which the environment confirms independently: its
projection diagnostics (projection frequency and norm, certification-zeroed
hours) are identically zero in all reported experiments.

\section{Experimental Setup}
\label{sec:p2setup}

\paragraph{Simulator and integration.}
All experiments run on the accepted SkillChain-Gym simulator from the
companion benchmark paper, imported read-only; no Paper~1 code was modified.
The default instance has two products, three skills, four workers, one
aggregate capacity pool, threshold $\theta_k = 0.6$, training gain
$\alpha^{\mathrm{train}} = 0.05$ per hour, forgetting
$\delta_k = 0.005$ per shift, and horizon $T = 60$ shifts. The only
configuration change relative to the benchmark tables is a wider observation
forecast window (16 shifts $\ge H_{\max}+1$) so receding-horizon controllers
can see announced shocks; \emph{every} policy --- MPC and baseline alike ---
receives the same widened observation, so observation parity holds within
every comparison (baseline numbers therefore differ from the benchmark
paper's window-3 tables). MPC plans both production and training with the
window, while the reused heuristics consume the lookahead only in their
training rule; their production rule is single-step by design.

\paragraph{Protocol lock.}
The primary configuration is SkillMPC with $\lambda_{\mathrm{gap}} = 25$ and
$H = 10$, chosen ex ante before the validation suite ran. Horizon and
$\lambda_{\mathrm{gap}}$ sweeps are sensitivity analyses, not model
selection; no headline claim is based on a best-of-sweep configuration. In
all artifacts, \texttt{solve\_status='heuristic'} marks Paper~1 baseline rows
(no solver involved); it is \emph{not} an MPC fallback. MPC fallbacks are
counted separately and are zero throughout.

\paragraph{Scenario families.}
The core suite uses five scenarios at the default training rate:
\texttt{no\_shock\_sanity}; \texttt{new\_product\_announced\_t60} (a product
activating mid-episode whose primary skill no worker initially holds,
visible in the forecast window beforehand);
\texttt{new\_product\_surprise\_t60} (activation shift randomized per seed
and hidden from the forecast until onset); \texttt{demand\_shock\_mid}
(temporary demand spike); and \texttt{absenteeism\_mid} (the two holders of
one skill absent for eight shifts). Around this core, the suite adds:
demand-forecast-quality modes on the announced shock (noisy, multiplicative
$\sigma = 0.15$; delayed by three shifts, emulated by truncating the
effective window; biased, visible forecast scaled by $0.8$);
availability-forecast modes on absenteeism (naive persistence; announced
absence, where the controller receives the exact absence window; noisy
announced, expecting half the lost hours); capacity-slack boundary sweeps on
the surprise shock (capacity $30, 31, 40$ against post-activation demand of
$30$ labor-hours, plus a scaled-demand check at $16{+}8$ demand with
capacity $24, 26$); training-rate sensitivity on the announced shock
(default $\alpha = 0.05$, moderate $0.025$, slow $0.0125$, with static
baselines recalibrated per rate); and two negative controls (a no-bottleneck
control with $\delta = 0$, and a near-infeasible control with post-activation
demand at the labor envelope). A final evidence pass adds three cells with
certification-event metrics: \texttt{recert\_demand\_shock},
\texttt{announced\_acquisition}, and \texttt{greenfield\_visible} (demand for
a skill held by no worker at $t{=}0$, forecast-visible from the first
shift).

\paragraph{Baselines and variants.}
ProductionOnlyMPC, MaintenanceMPC, and SkillMPC with
$\lambda_{\mathrm{gap}} \in \{0, 25, 100\}$ and $H \in \{3, 5, 10, 15\}$;
the static insurance plans (Static40 and Static60 at the default rate,
Static80 at the moderate rate, StaticSlow160 at the slow rate; the
uncalibrated Static40 is additionally kept in the slow cell, clearly
labeled, to document calibration fragility); and the water-filling
skill-gap heuristic (Section~\ref{sec:variants}).

\paragraph{Metrics.}
Per episode: total cost; service level and per-product service; total and
peak backlog; recovery rate, recovery time conditional on recovery, and
unrecovered counts; training hours (total and per skill); terminal new
certifications; within-episode \emph{recertifications} (upward
$\theta$-crossings of pairs certified at $t{=}0$ or previously acquired) and
\emph{greenfield acquisitions} (first crossing of pairs not certified at
$t{=}0$) --- counters added because the terminal metric is blind to
lapse-and-recover events; skill-bottleneck severity; and solver diagnostics
(status, mean solve time, fallback count, realized terminal gap) together
with the environment's projection diagnostics.

\paragraph{Statistical protocol.}
Twenty seeds per validation and final-evidence cell, paired by seed. Policy
comparisons report the per-seed win rate with an exact two-sided paired sign
test, and separately a seeded paired bootstrap (10{,}000 resamples)
percentile confidence interval on the mean cost difference with relative
effect sizes. The two criteria are reported side by side and are not
interchangeable: a comparison can show a favorable mean effect whose
bootstrap CI excludes zero (``ci-sig'') while its sign test is not
significant; we describe such cases as mean-effect evidence, not decisive
win-rate evidence. Episodes replay deterministically under fixed seeds: two
independent executions of each suite produce byte-identical CSVs once the
wall-clock solve-time column is excluded.

\paragraph{Artifacts.}
The validation suite comprises 87 aggregate cells and 1{,}740 seed-level
rows; the final evidence pass adds 18 cells and 360 rows. All artifacts,
per-seed sweeps, figures, and reproduction commands accompany the paper.

\section{Results}
\label{sec:p2results}

The results are organized by mechanism and regime rather than as a single
ranking. The summary finding is regime dependence: the predictive controller
is favored in the tested regimes when bottlenecks or labor shocks are
forecastable, while lean static insurance remains strong under surprise
shocks, near the demand--capacity boundary, and wherever insurance is cheap.
We report adverse regimes before favorable ones. Throughout, ``Primary'' is
the ex-ante locked configuration
(SkillMPC, $\lambda_{\mathrm{gap}} = 25$, $H = 10$); win rates come with
exact paired sign tests, and ``ci-sig'' marks paired-bootstrap confidence
intervals on the mean cost difference that exclude zero --- the two criteria
are reported separately and can disagree.

\subsection{Implementation and reproducibility}
\label{sec:res-impl}

The validation suite executes 67{,}200 MILP solves in closed loop with
\emph{zero} MPC fallbacks, zero NaNs, zero negative state metrics, and
environment projection diagnostics identically zero (every executed action
exactly feasible). Rows marked \texttt{solve\_status='heuristic'} are
Paper~1 baselines, not MPC failures. Median solve times range from
${\sim}4$\,ms ($H{=}3$) to ${\sim}94$\,ms ($H{=}15$), with a worst cell mean
of $0.57$\,s in the near-infeasible control. Two independent executions of
each suite produce byte-identical CSVs once the wall-clock
\texttt{solve\_ms\_mean} column is excluded.

\subsection{Attribution: what the skill machinery is worth}
\label{sec:res-attribution}

Table~\ref{tab:p2attrib} decomposes the controller's value through the
ablation chain ProductionOnly $\to$ Maintenance $\to$ Primary, which holds
the inventory-anticipation capability constant (all three share the same
inventory/backlog program and forecast) and varies only training
eligibility.

\begin{table}[t]
\centering
\caption{Attribution chain (20 seeds; mean cost; certification events per
episode from the final-evidence counters). On the new-product cells,
MaintenanceMPC equals ProductionOnlyMPC bit-for-bit on every seed.}
\label{tab:p2attrib}
\small
\begin{tabular}{lrrrrr}
\toprule
 & \multicolumn{2}{c}{demand\_shock\_mid} & \multicolumn{3}{c}{greenfield\_visible} \\
\cmidrule(lr){2-3}\cmidrule(lr){4-6}
Policy & cost & recerts & cost & greenfield certs & svc$_{p1}$ \\
\midrule
ProductionOnlyMPC & 21{,}235 & 0.0 & 138{,}854 & 0.0 & 0.00 \\
MaintenanceMPC    & 7{,}698  & 0.0 & 138{,}854 & 0.0 & 0.00 \\
Primary           & 3{,}709  & 1.0 & 2{,}225   & 2.0 & 0.999 \\
\bottomrule
\end{tabular}
\end{table}

On \texttt{demand\_shock\_mid} the chain is
$21{,}235 \to 7{,}698 \to 3{,}709$. The first leg ($77\%$ of the gap) is
certification \emph{maintenance}: without training, the production-only
controller loses both initially held certifications of the demanded skill to
forgetting mid-episode. The second leg ($23\%$) is \emph{not} new-skill
acquisition --- both policies train only the already-held skill and terminal
new-certification counts are zero for both --- but \emph{re-acquisition of
lapsed certifications}: the within-episode counters show Primary records
$1.0$ recertifications per episode on the demanded skill while
MaintenanceMPC records $0.0$, because once a certification lapses the
maintenance ablation can never recover it by construction. That Primary's
recertification count is $1.0$ while its \emph{terminal} new-certification
count is $0$ demonstrates directly that the terminal metric is blind to
lapse-and-recover events; the within-episode counters were added for this
reason. New-skill acquisition is exercised separately: on the new-product
cells (announced, surprise, and greenfield) MaintenanceMPC equals
ProductionOnlyMPC bit-for-bit on every seed --- neither can touch the
initially unheld skill --- so the entire gap to Primary there is acquisition
value. On \texttt{no\_shock\_sanity} the maintenance leg is the whole story
(Primary $1{,}832$ vs.\ Maintenance $1{,}833$, 18/20 ties, vs.\
ProductionOnly $16{,}735$), and in the $\delta = 0$ negative control the
chain degenerates: Primary equals ProductionOnlyMPC exactly, with zero
training hours --- the skill machinery contributes nothing when no skill
bottleneck exists.

\subsection{Where static insurance remains strong}
\label{sec:res-adverse}

\begin{table}[t]
\centering
\caption{Core T$=60$ results (20 seeds): mean cost $\pm$ s.d. Sign-test
notes in text. PO = ProductionOnlyMPC, Maint = MaintenanceMPC,
WF = WaterFillingSkillGap.}
\label{tab:p2core}
\small
\setlength{\tabcolsep}{4pt}
\begin{tabular}{lrrrrr}
\toprule
Policy & no\_shock & announced & surprise & absenteeism & demand\_shock \\
\midrule
Primary   & 1{,}832 $\pm$ 81 & 1{,}992 $\pm$ 82 & 2{,}110 $\pm$ 168 & 6{,}675 $\pm$ 584 & 3{,}709 $\pm$ 471 \\
SkillMPC $\lambda{=}0$ & 1{,}832 $\pm$ 81 & 1{,}991 $\pm$ 83 & 2{,}126 $\pm$ 181 & 6{,}681 $\pm$ 582 & 3{,}719 $\pm$ 471 \\
PO  & 16{,}735 $\pm$ 685 & 67{,}861 $\pm$ 1{,}060 & 44{,}241 $\pm$ 33{,}517 & 36{,}239 $\pm$ 1{,}157 & 21{,}235 $\pm$ 1{,}067 \\
Maint & 1{,}833 $\pm$ 81 & 67{,}861 $\pm$ 1{,}060 & 44{,}241 $\pm$ 33{,}517 & 13{,}595 $\pm$ 4{,}363 & 7{,}698 $\pm$ 3{,}031 \\
Static40  & 3{,}611 $\pm$ 684 & 2{,}108 $\pm$ 130 & \textbf{1{,}981 $\pm$ 173} & 6{,}652 $\pm$ 718 & 8{,}802 $\pm$ 830 \\
Static60  & 2{,}246 $\pm$ 145 & 2{,}133 $\pm$ 115 & 2{,}005 $\pm$ 159 & \textbf{4{,}915 $\pm$ 666} & 6{,}991 $\pm$ 795 \\
WF        & 1{,}893 $\pm$ 170 & 2{,}263 $\pm$ 242 & 2{,}658 $\pm$ 384 & 7{,}238 $\pm$ 906 & 7{,}348 $\pm$ 1{,}102 \\
\bottomrule
\end{tabular}
\end{table}

\paragraph{Surprise shocks.} Lean static insurance wins the surprise
new-product regime on cost: Static40 at $1{,}981$ beats Primary at $2{,}110$
on 15 of 20 paired seeds (sign $p = 0.041$, ci-sig, $+6.5\%$ relative), and
Static60 likewise. The controller cannot anticipate a hidden activation any
better than the heuristics can; its advantage over the reactive
water-filling heuristic ($2{,}658$, Primary wins 20--0) comes from cleaner
reaction, not foresight.

\paragraph{Absenteeism without availability forecasts.} Under the default
persistence assumption, Static60 at $4{,}915$ beats Primary at $6{,}675$ on
20 of 20 seeds ($+36\%$): the static plan's cross-trained backups are
pre-bought redundancy, while the controller only reacts once workers are
already absent. Primary ties Static40 (10--10).

\paragraph{The demand--capacity boundary.} Table~\ref{tab:p2boundary} sweeps
post-activation capacity slack on the surprise shock. At zero slack
(capacity $30$ against $30$ demanded hours) any reaction transient is
structurally hard to recover: Primary recovers in only $45\%$ of episodes
(11/20 unrecovered) and loses to Static40 20--0 ($3{,}305$ vs.\ $2{,}089$,
$+58\%$); the reactive heuristic recovers in $5\%$. One unit of slack
restores Primary's recovery to $95\%$ but Static40 still wins 17--3; even at
generous slack (capacity 40) Static40 retains a small significant edge
(16--4, $+7\%$). The scaled-demand check ($16{+}8$ demand) reproduces the
qualitative pattern at a different absolute scale, with static favored
throughout because cheap pre-shock labor makes its insurance nearly free.

\begin{table}[t]
\centering
\caption{Capacity-slack boundary on the surprise shock (20 seeds): mean
cost, recovery rate, and unrecovered episodes. Post-activation demand is 30
labor-hours (default scale) and 24 (scaled).}
\label{tab:p2boundary}
\small
\begin{tabular}{llrrr}
\toprule
Capacity (slack) & Policy & cost & recovery & unrecovered \\
\midrule
30 (0) & Primary  & 3{,}305 & 0.45 & 11 \\
       & Static40 & 2{,}089 & 0.95 & 1 \\
       & WF       & 5{,}426 & 0.05 & 19 \\
31 (1) & Primary  & 2{,}265 & 0.95 & 1 \\
       & Static40 & 1{,}981 & 1.00 & 0 \\
40 (8+, labor-bound) & Primary  & 2{,}121 & 1.00 & 0 \\
       & Static40 & 1{,}981 & 1.00 & 0 \\
\midrule
24 (0, scaled) & Primary & 2{,}371 & 0.90 & 2 \\
       & Static40 & 1{,}691 & 0.95 & 1 \\
26 (2, scaled) & Primary & 1{,}703 & 1.00 & 0 \\
       & Static40 & 1{,}621 & 1.00 & 0 \\
\bottomrule
\end{tabular}
\end{table}

\paragraph{Near-infeasible demand.} When post-activation demand sits at the
labor envelope, no policy can fully serve, and the cheaper insurance wins:
Static40 $2{,}484$ beats Primary $3{,}411$ on 20 of 20 seeds. Predictive
control does not rescue a structurally overloaded system.

\subsection{Where forecastability favors the controller}
\label{sec:res-favorable}

\paragraph{Announced bottlenecks.} On the announced new-product shock,
Primary at $1{,}992$ shows a favorable mean effect against Static40 at
$2{,}108$ (ci-sig, $-5.5\%$ relative), but wins only 13 of 20 paired seeds
(sign $p = 0.26$): we report this as mean-effect evidence, not decisive
win-rate evidence. Against Static60 the advantage is sign-significant
(17--3, $p = 0.0026$) and against the water-filling heuristic decisive
(20--0, $-12\%$). The controller achieves this with $13.4$ training hours
versus the static plans' $40$--$60$: it buys only the certifications the
forecast justifies, just in time.

\paragraph{Demand shocks.} Primary at $3{,}709$ beats every baseline on 20
of 20 seeds (Static40 $8{,}802$, $-58\%$; Static60 $6{,}991$, $-47\%$;
water-filling $7{,}348$, $-50\%$): receding-horizon inventory pre-building
ahead of the announced spike, combined with certification maintenance, is a
capability none of the open-loop or single-step baselines has. The same
holds in \texttt{no\_shock\_sanity} against the static plans (20--0), where
Primary additionally ties the water-filling heuristic (9--11, n.s.).

\paragraph{Forecastable labor shocks.} The sharpest reversal is the
availability experiment. With the default persistence assumption the
controller loses absenteeism to Static60 (above); given an \emph{announced}
absence window, Primary drops from $6{,}675$ to $2{,}780$ --- beating its
own persistence variant 20--0 ($-58\%$) and Static60 ($4{,}915$) 20--0
($-43\%$). A noisy announcement (half the lost hours expected) recovers
part of the value ($5{,}668$). Forecastable labor shocks move the
absenteeism regime from static insurance to predictive control; surprise
labor shocks do not.

\subsection{Terminal skill-bottleneck value: a dose--response in training
lag}
\label{sec:res-terminal}

\begin{figure}[t]
\centering
\includegraphics[width=0.85\linewidth]{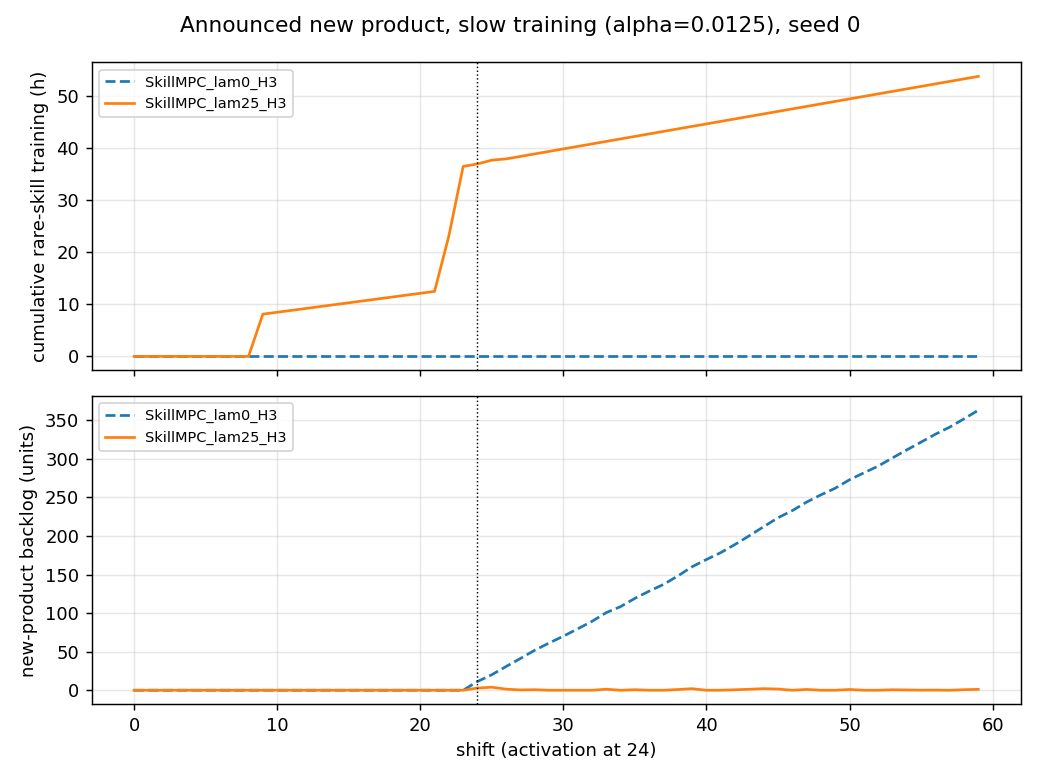}
\caption{Announced new product under slow training ($\alpha = 0.0125$,
seed~0, $H = 3$): cumulative rare-skill training (top) and new-product
backlog (bottom). Without the terminal penalty the controller never trains
and backlog grows without bound; with $\lambda_{\mathrm{gap}} = 25$ training
starts the first shift the activation becomes visible (shift~9) and backlog
stays near zero.}
\label{fig:p2slowtrain}
\end{figure}

The terminal penalty's effect depends on the relation between training lag
and horizon, and the three training-rate regimes must not be pooled. At the
\emph{default} rate (certification within one shift), just-in-time training
is optimal in the predictor and the penalty has no cost effect at $H = 10$:
$\lambda_{\mathrm{gap}} = 25$ vs.\ $0$ is 8--8 with 4 ties on the announced
shock, although the realized terminal-gap diagnostic drops from $28.5$ to
$0.0$ --- the penalty changes the predicted plan, not the executed cost. At
the \emph{moderate} rate ($\alpha = 0.025$, lag ${\approx}2$ shifts),
$\lambda_{\mathrm{gap}} = 25$ at $H = 3$ costs $2{,}027$ versus $3{,}209$
without the penalty. At the \emph{slow} rate ($\alpha = 0.0125$, lag 2--4
shifts), the penalty becomes decisive for short horizons: without it the
$H = 3$ controller never trains and matches production-only behavior
($67{,}861$, new-product service $0$, while its own gap diagnostic sits at
$28.5$); with it, training starts at shift 9 --- the first shift the
activation enters the forecast window --- and cost falls to $2{,}035$
($-97\%$, 20--0; Figure~\ref{fig:p2slowtrain}). At $H = 5$ the no-terminal
controller trains late and partially recovers ($11{,}231$); the with-terminal
variants are indistinguishable from the $H = 10$ primary. Against fairly
recalibrated static plans the controller retains a decisive advantage on
this announced shock: Primary beats StaticSlow160 ($2{,}037$ vs.\ $4{,}722$,
20--0) and Static80 at the moderate rate ($2{,}003$ vs.\ $2{,}756$, 20--0),
while the \emph{uncalibrated} Static40 fails completely at the slow rate
($67{,}945$, labeled calibration fragility).

\subsection{Forecast quality and horizon sensitivity}
\label{sec:res-quality}

On the announced shock, degrading the forecast degrades the controller in
an interpretable order: perfect $1{,}992$; delayed-by-3 $1{,}992$ (exactly
zero effect on all 20 seeds); noisy $3{,}002$ ($+51\%$, 0--20 against
perfect); biased ($0.8\times$) $5{,}189$ ($+160\%$, 0--20). The delay
result is a property, not an anomaly: a delay harms the controller only
once remaining visibility falls below the training lag plus one shift, and
at the default rate (lag one shift) visibility of even a single shift
suffices; under slow training, larger delays bind correspondingly earlier.
Horizon sensitivity at the locked $\lambda_{\mathrm{gap}} = 25$ is mild in
the default-rate regime ($H = 3/5/10/15$: $2{,}017/2{,}014/1{,}992/1{,}995$
announced; $2{,}133/2{,}108/2{,}110/2{,}098$ surprise), consistent with the
terminal value substituting for horizon length only when training lag binds.

\subsection{Greenfield acquisition}
\label{sec:res-greenfield}

In \texttt{greenfield\_visible} (no worker holds the required skill at
$t{=}0$; demand visible from the first shift), Primary acquires $2.0$
greenfield certifications per episode with $14$ training hours and reaches
cost $2{,}225$. The comparison with Static40 ($2{,}343$) is favorable on
the mean with a bootstrap CI excluding zero ($-118$, CI $[-203, -38]$) but
\emph{not} sign-test significant (13--7, $p = 0.26$); we phrase it
conservatively as mean-effect evidence. Against Static60 the advantage is
sign-significant (15--5, $p = 0.041$) and against water-filling clear
(18--2, $p = 4\times10^{-4}$). MaintenanceMPC and ProductionOnlyMPC, which
cannot acquire the skill, match one another exactly ($138{,}854$;
new-product service $0$; 20--0). Together with
Sections~\ref{sec:res-attribution}--\ref{sec:res-terminal}, this completes
the mechanism picture: maintenance, re-acquisition, and greenfield
acquisition are separated, each with direct metric support.

\section{Discussion and Limitations}
\label{sec:p2discussion}

\paragraph{Reading the regime map conservatively.}
The headline of Section~\ref{sec:p2results} is not that predictive control
wins, but \emph{where} it wins, and the adverse regimes come first. Lean
static cross-training remains strong --- often sign-significantly better
than the locked primary controller --- under surprise shocks (Static40 wins
15--5), under absenteeism when the controller can only assume availability
persistence (Static60 wins 20--0), throughout the demand--capacity boundary
sweep (including 20--0 at zero slack, where reaction transients are
structurally hard to recover), and in the near-infeasible control where no
policy can fully serve and cheaper insurance wins 20--0. The $\delta = 0$
negative control adds the cleanest null: with no forgetting and no skill
bottleneck, the skill-aware controller reproduces the production-only
controller exactly, with zero training hours. Any claim about
skill-constrained MPC must survive these cells, and ours is correspondingly
narrow: predictive control is useful in the tested regimes when skill or
labor bottlenecks are forecastable early enough for training to complete.

\paragraph{Favorable regimes.}
Where the future is visible, the controller converts visibility into lean,
just-in-time capability decisions. On forecast-visible bottlenecks it buys
only the certifications the forecast justifies ($13$--$14$ hours of training
against the static plans' $40$--$60$), with a favorable mean effect against
the leanest static plan (ci-sig but not sign-significant, 13--7) and
sign-significant advantages beyond it. On demand shocks it combines
inventory pre-building with certification maintenance, a joint capability no
open-loop or single-step baseline has (20--0 against all). The sharpest
reversal is informational: given an announced absence window, the same
controller that loses absenteeism under persistence beats Static60 20--0 at
$-43\%$ cost. Forecastable labor shocks move a regime that Paper~1 assigned
to static insurance into the predictive-control column; hidden labor shocks
do not.

\paragraph{What the attribution chain establishes.}
The ProductionOnly $\to$ Maintenance $\to$ Primary chain, with
inventory-anticipation held constant, separates three mechanisms with direct
metric support. Certification \emph{maintenance} against forgetting is the
largest leg wherever skills are initially held (77\% of the demand-shock
gap; the entire no-shock gap). \emph{Re-acquisition of lapsed
certifications} is visible only through the within-episode recertification
counters (Primary $1.0$ vs.\ Maintenance $0.0$ per episode on the demanded
skill) and is provably invisible to terminal certification counts.
\emph{Greenfield acquisition} is exercised where no worker holds the
required skill: the maintenance ablation collapses onto production-only
bit-for-bit, and the full controller acquires exactly the certifications the
forecast justifies. Per-skill training hours confirm that each mechanism
operates on the expected skill and no other.

\paragraph{Forecast privilege, not magic adaptivity.}
The controller's advantages are announcement and forecast leverage, made
explicit rather than hidden. Degrading the forecast degrades the controller
in an interpretable order (noisy $+51\%$, biased $+160\%$), a delayed
forecast is harmless exactly until remaining visibility falls below the
training lag, and hiding the shock entirely (surprise) hands the regime back
to static insurance. We consider this transparency a feature of the
evaluation design: every policy receives the same observation, the surprise
scenarios show what happens when privilege is absent, and the
availability-forecast experiment quantifies precisely how much of the
absenteeism regime is purchasable with better labor forecasts.

\paragraph{Terminal skill-bottleneck value.}
The terminal penalty matters exactly where theory predicts: when the
training lag approaches or exceeds the horizon. At the default training rate
(one-shift lag) just-in-time training is optimal in the predictor and
$\lambda_{\mathrm{gap}}$ has no cost effect at $H{=}10$ --- the penalty
closes the predicted terminal gap without changing executed cost. At the
moderate and slow rates the no-terminal short-horizon controller trains too
late or not at all (up to $-97\%$ cost difference at the slow rate, where
the no-terminal controller sits on a visible gap diagnostic of $28.5$ and
never pays for it). The three regimes are reported separately and should
never be pooled; the terminal value is an interpretable cost-to-go proxy,
ablated at the same three $\lambda_{\mathrm{gap}}$ values everywhere rather
than tuned per scenario.

\paragraph{Limitations.}
The simulator is stylized by design: linear training gain, geometric
forgetting, hard threshold certification, and a small synthetic instance
(two products, three skills, four workers, one site). Procurement and
material flows, supplier delays, multi-echelon networks, job-shop routing,
soft productivity, and learning-by-doing are excluded; no real data are
used, and no demographic or broader fairness claims are made. Forecast
modes are simple transformations (noise, delay, bias, persistence) rather
than learned forecasters, and availability announcements are exact windows.
Solver scalability is demonstrated only at this scale (milliseconds per
solve; sub-second in the hardest cell) --- larger instances with more
workers, skills, or binary structure remain untested. The static insurance
baselines are favorable by construction (their plan structure encodes which
skills can become critical, and they are recalibrated per training rate),
which strengthens the adverse cells but also means the favorable comparisons
are against strong, informed opponents rather than strawmen. Finally, the
statistical protocol is paired over twenty seeds; borderline comparisons
(announced and greenfield against Static40) are reported as mean-effect
evidence with confidence intervals, not as decisive win-rate results.

\section{Conclusion}
\label{sec:p2conclusion}

We formulated and evaluated a closed-loop skill-constrained model predictive
controller for a single-site production-inventory system in which worker
skills are dynamic states, certification is a hard threshold, and training is
a control action that competes with production for the same worker hours. At
every shift the controller solves a finite-horizon mixed-integer program with
binary predicted certification and an interpretable terminal
skill-bottleneck penalty, applies only the first-period action, and replans;
all 67{,}200 solves in the validation suite completed without fallback, the
executed actions were exactly feasible throughout, and every artifact
replays deterministically under fixed seeds.

The central empirical result is regime dependence, established against
strong, deliberately favorable static-insurance baselines and a
locked-in-advance primary configuration. Predictive control is useful when
skill or labor bottlenecks are forecastable early enough for training to
complete: forecast-visible new-product bottlenecks, announced demand shocks
where inventory anticipation combines with certification maintenance,
announced absence windows, and slow-training regimes where the terminal
skill-bottleneck value prevents short-horizon controllers from ignoring
visible future gaps. Static insurance remains hard to beat when shocks are
hidden, when reaction transients are structurally unrecoverable near the
demand--capacity boundary, and wherever pre-shock slack makes insurance
cheap. No policy class dominates across regimes. The attribution chain
separates the controller's value into certification maintenance,
re-acquisition of lapsed certifications, and greenfield skill acquisition,
each supported by within-episode certification-event counters and per-skill
training evidence rather than terminal summaries alone.

Natural next steps follow the same discipline. Richer skill structures and
larger instances would test whether the regime boundaries persist when no
static plan can pre-train every contingency and when solver scale begins to
bind; stochastic or scenario-based receding-horizon formulations could price
hidden shocks that the present certainty-equivalent controller cannot
anticipate; learned or public-data-calibrated demand and absence forecasts
would replace the stylized forecast modes; and integration with the released
SkillChain-Gym benchmark of the companion paper would let other controllers
and learned policies be evaluated under the same scenarios, metrics, and
statistical protocol.

\section*{Funding}
This research received no external funding.

\bibliographystyle{plain}
\bibliography{references}

\end{document}